\documentclass[aapm,mph,preprint,graphicx]{revtex4-2}
\draft
\usepackage[mathlines]{lineno}% Enable numbering of text and display math 

\usepackage{ragged2e}
\usepackage{caption}
\usepackage{subcaption}
\usepackage{graphicx}
\usepackage{lipsum}
\usepackage{amsmath}
\usepackage{indentfirst}
\usepackage{multirow}
\usepackage{xcolor}
\begin{document}

\title{Changes in Visual Attention Patterns for Detection Tasks due to Dependencies on Signal and Background Spatial Frequencies} %Title of your paper .\\

\justify
\author{Amar Kavuri$^1$}
\author{Howard C. Gifford$^1$}
\author{Mini Das$^{1,2,3}$}

\email[Email: ]{mdas@uh.edu}
\thanks{Author to whom correspondence should be addressed.}
%\homepage[]{Your web page}
%\altaffiliation{}
\affiliation{$^1$Department of Biomedical Engineering, University of Houston, Houston,TX-77204,USA}

\affiliation{$^2$Department of Physics, University of Houston, Houston,TX-77204,USA}

\affiliation{$^3$Department of Electrical and Computer Engineering, University of Houston, Houston,TX-77204,USA}

\date{\today}
\justify
\begin{abstract}

We aim to investigate the impact of image and signal properties on visual attention mechanisms during a signal detection task in digital images. The application of insight yielded from this work spans many areas of digital imaging where signal or pattern recognition is involved in complex heterogenous background. We will use simulated  tomographic  breast images as the platform to investigate this question. While radiologists are highly effective at analyzing medical images to detect and diagnose diseases, misdiagnosis still occurs.  These errors can stem from multiple sources such as the inherent properties of medical images and the accompanying signal characteristics play a pivotal role in visual search strategy while locating and identifying the disease. 
We selected digital breast tomosynthesis (DBT) images as a sample medical images with different breast densities and structures using digital breast phantoms (Bakic and XCAT). Two types of lesions (with distinct spatial frequency properties) were randomly inserted in the phantoms during projections to generate abnormal cases. Six non-radiologist human observers participated in observer study designed for a locating and detection of an 3-mm sphere lesion and 6-mm spicule lesion in reconstructed in-plane DBT slices. We collected eye-gaze data to estimate gaze metrics and to examine differences in visual attention mechanisms.
Gaze analysis revealed that diagnostic response times were significantly longer in Bakic phantoms and high-density tissue backgrounds compared to XCAT breast phantoms (p < 0.05) and lower-density backgrounds (p < 0.01) respectively, indicating increased perceptual difficulty in anatomically complex scenes. Observers made longer fixations on spiculated lesions compared to spherical lesions (p < 0.01), suggesting that lesion morphology modulates visual attention allocation. Across all attention stages—search, recognition, and decision—the small spherical lesion required higher contrast for successful detection, especially in the Bakic background where anatomical noise severely reduced its visibility. Error analysis showed that the majority of misdiagnosed cases (25\%) occurred when the lesion was not visible, followed by decision-stage errors (11\%), where the signal was seen but incorrectly judged.

Conclusions: Diagnostic performance in complex visual environments is strongly constrained by later perceptual stages, with decision failures accounting for the largest proportion of errors. Lesion detectability is jointly influenced by both target morphology and background complexity, revealing a critical interaction between local signal features and global anatomical noise. Increased fixation duration on spiculated lesions suggests that visual attention is differentially engaged depending on lesion structure, potentially aiding recognition. These findings highlight the importance of perceptually informed design and training of computer aided diagnosis systems.

%reader training strategies that target specific stages of visual processing under diagnostic uncertainty.
%establish minimum fixation duration (times) to locate the signal accurately with respect to signal contrast. expert may not need longer time, signal size and type may change this duration amount. 

\end{abstract}

\pacs{}
\keywords{ eyetracking; visual attention;  signal detection; spatial frequencies }

\maketitle

\section{INTRODUCTION}\label{intro}

Disease detection and diagnosis frequently involve the analysis of medical images. For example, in cancer screening, anomalous growths are primarily identified using medical images. In a single day, clinicians can be tasked with analyzing a large volume of images in search of a disease\cite{nakajima2008radiologist,mcdonald2015effects}. Although experienced clinicians can be highly effective at this task, misdiagnosis still occurs in some cases\cite{brady2017error}. Two major factors can contribute to these errors. The first one is the limitations of the image acquisition system which can hinder the visibility of anomalies\cite{timberg2012visibility}. The second factor is related to the limitations of the human eye-brain system, and it can result in misses even when an anomaly (e.g., a cancerous growth) is visible\cite{brady1994colorectal,brady2017error}. Measuring the eye movements of observers during these tasks can help us expand our understanding of how medical images are analyzed and, in turn, explain the reasons for errors\cite{kundel1978visual}.

% In practice, however, overlapping anatomy and variable tissue patterns continue to make visual search difficult. For example, dense fibroglandular tissue in mammograms degrades 2D detection, and while DBT helps by spreading tissue over slices, dense backgrounds still create strong visual clutter. Understanding how the statistical structure of these backgrounds affects human search strategies is therefore critical to improving detection performance.

In recent years, gaze analysis using eye-tracking systems has gained significant interest in medical image research\cite{leveque2018state}. The analysis of gaze data has revealed important findings about human visual attention mechanisms during diagnostic tasks. Carmody et al. \cite{carmody1981finding} demonstrated that diagnostic errors are influenced by both radiologists' visual scanning strategies and nodule visibility in chest X-ray films. Kundel et al. \cite{kundel1978visual} found that radiologists sometimes failed to report cancers as positive even after fixating on the cancer locations, suggesting that false-negative errors also stem from recognition failure. Based on eye-tracking studies, these false-negative errors have been classified into three categories\cite{kundel1978visual,brunye2019review,wu2019eye}: search errors (never fixating on the cancer location), recognition errors (fixating once but failing to recognize the cancer), and decision errors (fixating multiple times or for longer durations but still failing to report positively). 
These errors can stem from multiple sources beyond human factors alone. Among these, the inherent properties of medical images and the accompanying signal characteristics play a pivotal role in shaping search strategy\cite{aizenman2017comparing,mousa2014mammographic}.

Image quality attributes such as contrast, spatial resolution, noise, artifacts, overlapping anatomy and variable tissue patterns directly affect the visibility and detectability of anatomical structures and pathological features\cite{burgess2007signal,abbey2014observer,conrey2009pattern}, influencing the likelihood of perceptual and cognitive errors during image interpretation. Furthermore, signal properties including signal contrast and shape modulate the clarity and reliability of the visual information available to clinicians. For instance, Burgess et al.\cite{burgess2007signal} showed that spectral properties of filtered noise backgrounds significantly impact signal detection performance—larger signals are harder to detect in coarser texture backgrounds, while smaller signals are harder to detect in finer texture backgrounds. Abbey and Eckstein \cite{abbey2014observer} extended the study for search and localization tasks in different background noises, demonstrating that observers might use different target frequency channels based on task type (location known versus unknown), even when viewing signals and backgrounds with identical properties. Despite these insights, the influence of factors such as heterogeneity in patient anatomies and signal type on attention mechanisms in tomographic images remained vastly underexplored. Understanding how these image and signal parameters contribute to different types of diagnostic errors is essential to develop targeted strategies that enhance image acquisition protocols, optimize processing algorithms, and improve observer performance. Therefore, this research hypothesis focuses on investigating the influence of image and signal properties on the prevalence and nature of diagnostic error types, with the aim of reducing errors such as search and recognition mistakes, ultimately improving diagnostic accuracy and patient care quality. 

%Eye-tracking studies have also revealed that radiologists can fixate on abnormality regions within the first second of viewing an image\cite{kundel2008using}, demonstrating a rapid "gist" or holistic processing stage that influences subsequent analysis.

 To accomplish this, we selected digital breast tomosynthesis (DBT) images as a sample medical images with different breast densities and structures. We conducted human observer studies on in-plane DBT images for search and lesion localization tasks and collected eye-gaze data to estimate gaze metrics and examine differences in visual attention mechanisms.

%Few studies explored how background texture influences different attentional mechanisms in medical imaging. Here, we choose phantoms that were generated using different processes to emulate different texture backgrounds. a critical gap remains in understanding how background structural variations specifically influence the different attentional mechanisms involved in medical image interpretation.
\section{Methods}
The study methods are summarized below. First, the selected breast phantoms with different densities and structres and signal are presented. Second, the simulation methodology is described to generate DBT images using the phantoms and signals. Next, the experimental setup to conduct human observer study by collecting eye gaze data is described. Finally, the metrics used to estimate human observer performance and to analyze eye tracking data are presented.

\subsection{Stimuli preparation}
 We selected two types of digital breast phantoms to simulate the breast background structures. The first type was based on the three-dimensional anthropomorphic breast models generated by Bakic’s group at the university of Pennsylvania \cite{bakic2002mammogram} and for the second type we used XCAT breast phantoms created at Carl E. Ravin Advanced Imaging Laboratories (RAILabs) at Duke university using compressed volumes of patient breast CT data \cite{erickson2016population}. For our analysis, we selected six digital phantoms (5 cm thick) of each type. Three phantoms have approximately 25\% volumetric glandular fraction (VGF) while other three phantoms have approximately 50\% VGF. This wide gap in breast densities was chosen to give a wide difference in background complexity for the detection study. 

\begin{figure}[ht!]
\centering\includegraphics[width=7cm]{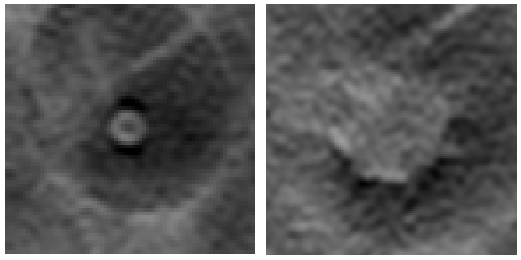}
\caption{\label{fig:lesions}Sample regions with spherical (left) and speculated (right) lesions.}
\end{figure}

We simulated two types of lesions to use as positive indications in our DBT simulations. The first kind were spherical lesions measuring 3-mm in diameter. The second type of lesions were speculated lesions with a mean mass diameter of 6-mm, which we generated with the methodology described in ref\cite{de2015computational}. Figure \ref{fig:lesions} shows the sample regions of DBT slices with two types of lesions.

The DBT images used in this study were generated using a serial cascade model. We provide a brief description of the simulation platform in this work, while more details can be found in some of our previous work \cite{vedula2003computer, kavuri2020relative}.
We simulate x-ray transmission through the breast using the  Siddon’s ray tracing method. 
%Quantum noise in each projection was modeled based on Poisson distribution. 
Abnormal cases were generated by digitally inserting simulated lesions at randomly selected locations within volumetric phantom prior to forward projection. To reconstruct the three-dimensional images, we used Feldkamp filtered back projection. For abnormal cases, 1-mm axial slices were extracted at the depth corresponding to the center coordinates of the inserted lesions, ensuring that the lesion was fully visible. Normal cases were generated by extracting 1-mm slices at random depths from reconstructed volumes without lesions.
Because lesion placement occurred in anatomically diverse locations, the surrounding structural context varied, leading to differences in local contrast and lesion conspicuity. Signal contrast was quantified as the difference between the mean pixel intensity of the lesion region and the mean intensity of adjacent background tissue within a 1-mm margin.

\subsection{Apparatus}
\begin{figure}[ht!]
         \centering\includegraphics[width=0.7\textwidth]{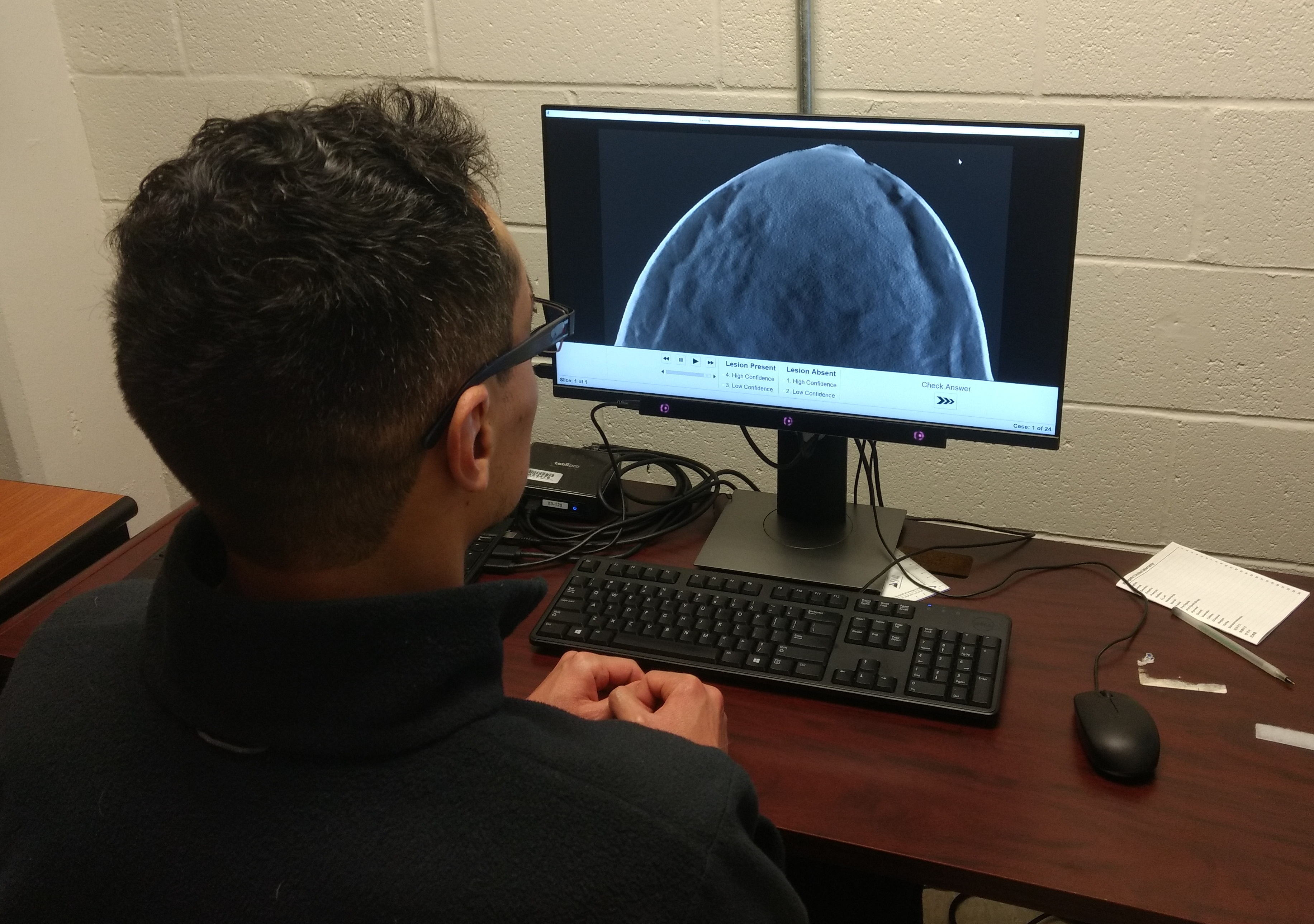}
        \caption{\label{fig:ET_setup}Setup of remote eye-tracking device attached to the monitor and the setup of in-house built graphical user interface (GUI).}
\end{figure}

The eye-tracker hardware used for this study was Tobii pro X3-120 with EPU. This eye tracker model is a screen-based tool that allows users of free head movement within the limits. A python-based Tobii pro SDK(v1.7) was installed on Windows based computer to interact with the hardware. An in-house graphical user interface (GUI), developed in python, was used to present the images and provide a way to annotate the location of lesion and score confidence levels. The images were displayed on a standard Dell 23.8 inch LCD monitor with a resolution of 1920 $\times$ 1080 as shown in figure \ref{fig:ET_setup}. The setup allows the experimenter to conduct both training and testing sessions.
% During the experiments, the images presented to observers were randomly selected from a databank specific to the aims of the study.  The images were displayed to the user in a random order.  In the training session, truth information was shown as feedback to the observer.

In addition, eye-tracker functionalities such as initiating the eye-tracking system, calibration, collecting and storing the gaze logs along with user selections were also implemented. The first step in post processing the gaze data is reducing the noise and estimating the fixation locations, fixation duration, saccade length, and saccade duration. 
Fixations were identified using the Tobii I-VT (Velocity-Threshold Identification) filter. Gaze samples were classified as fixations when gaze velocity was less than $30^\circ/s$  for a minimum duration of 60 ms, following the default parameters recommended in the Tobii I-VT filter white paper\cite{olsen2012tobii,tobii2012determining}. These values were determined by Tobii to provide stable and accurate fixation detection across a wide range of eye-tracking conditions. Adjacent gaze events were merged into a single fixation if they occurred within 75 ms and were separated by less than $0.5^\circ$, reducing fixation fragmentation caused by noise or brief velocity fluctuations.%For post processing, an I-VT filter was used based on Tobii pro white paper\cite{olsen2012tobii}. We define a gaze event as fixation when the gaze velocity is less than $30^\circ/s$  for a minimum duration of 60 ms. Two adjuscent gaze events were merged as one when they occur within 75 ms and $0.5^\circ$.

\subsection{Procedure}
Six non-radiologists took part in the human observer experiments to detect and localize the abnormality. Physicists and engineers who participated as observers had the same level of experience in reading simulated images. The method used to conduct the human observer experiments was same as that described in our prior work Ref.\cite{das2015examining,das2011comparison,lau2013towards,gifford2016visual}, and it is briefly summarized here. In our human observer studies, in-plane DBT slices of the four sets were evaluated. There were 96 images per set (48 pairs of abnormal/normal images), divided into 72 study images and 24 training images. Each set included an initial training session followed by a test session. In the training session, following the observer’s localization of lesion, the ground truth was displayed (either there was no lesion present or location of the lesion indicated by a square). Each observer read the four image sets. The reading order of images in each set was randomized for each observer. Room lighting and display were kept constant, but observers were permitted to vary the viewing distance and angle. Observers were asked to select the lesion location and allocate a confidence rating from 1 to 4 where 1 indicates high confidence lesion absent, 2 indicates low confidence lesion absent, 3 indicates low confidence lesion present, and 4 indicates high confidence lesion present.
% The performance of observers was quantified with area under LROC curves (AUC). The estimate of AUC for a given observer and protocol was obtained with a Wilcoxon-based nonparametric ranking method.
%Localizations were considered as correct when the observer selected the location within a 4-mm radius of lesion center.

A second set of human observer studies were conducted with similar setting as described above except that lesion location was indicated with a square box around the lesion for lesion present images and at random location for lesion absent images. The goal of the second set was to estimate the lesion visibility, hence only one of the six observers rated the images for the visibility of lesion. A lesion was categorized as not visible when the visibility rating is equal or lower than the $90^{th}$ percentile of visibility ratings of the lesion absent images.
\subsection{Data processing}
In literature, gaze patterns were shown to have relation with perceptual and cognitive processes and revealed many interesting findings\cite{brunye2019review,kundel1978visual,kundel2008using,voisin2013investigating}.
Using the collected eye tracking data, we computed five gaze metrics to characterize the gaze patterns. Namely, we found the time taken to diagnose each image (total time), total  number of fixations made on each image, time taken to first fixate on the lesion region (first hit time), number of fixations on lesion, and lesion dwell time.

In addition to standard gaze metrics, a series of error signatures were estimated to characterize underlying visual attention mechanisms during lesion detection. A case was labeled as a miss or false negative (FN) when the observer failed to correctly localize a lesion. FN errors were further classified into four distinct categories based on gaze behavior and observer feedback: (i) not visible, as rated by an observer; (ii) search error, defined as the absence of any fixation on the visible lesion; (iii) recognition error, identified when the observer fixated on the visible lesion only once with a fixation duration of less than 1 second; and (iv) decision error, characterized by multiple fixations on the lesion or a cumulative fixation time of at least 1 second without correct localization.

The images were then ordered to reflect these stages of visual attention, ranging from cases where the lesion was not visible, to lesions that were visible but never fixated or searched, lesions that were fixated but not recognized, lesions that were recognized but not localized, and finally lesions that were successfully localized. To quantitatively model performance at each stage, Gaussian cumulative distribution functions (CDFs) were fit to estimate the probability of success as a function of signal contrast. Specifically, the probability of lesion visibility was estimated by fitting a Gaussian CDF to the classes of signal not visible versus signal visible. The probability of successful search was estimated by grouping cases where the signal was either not visible or not fixated, against those where lesion was fixated atleast once. Similarly, the probability of successful recognition was modeled by combining the classes of not visible, not fixated, and not recognized, versus atleast recognized. Finally, the probability of successfully deciding to localize the lesion was estimated by contrasting all failure categories against those in which the lesion was localized. For each stage of visual attention, the signal contrast required to achieve an 80\% probability of success was derived from the corresponding Gaussian CDF fit, providing a quantitative measure of contrast sensitivity throughout the lesion detection process.

%Statistical analyses were carried out using MATLAB2018a (The MathWorks, Inc.) with t-test.

\section{Results}

\subsection{Gaze metrics}

\begin{figure}[htbp]
     \centering
     \begin{subfigure}[b]{0.6\textwidth}
         \centering\includegraphics[width=\textwidth]{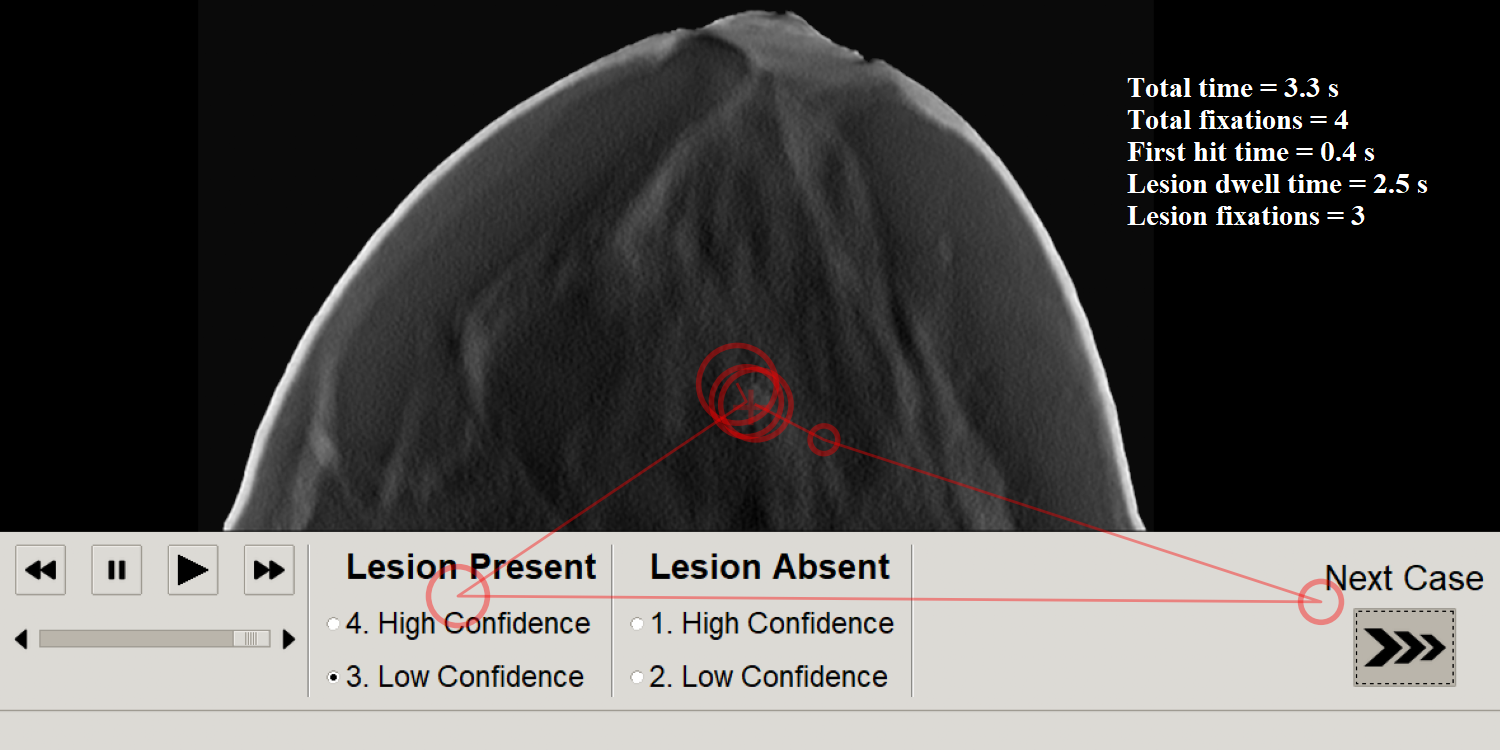}
         \caption{\label{fig:xcat}25\% dense XCAT breast background}
     \end{subfigure}
\vskip\baselineskip
     \begin{subfigure}[b]{0.6\textwidth}
         \centering\includegraphics[width=\textwidth]{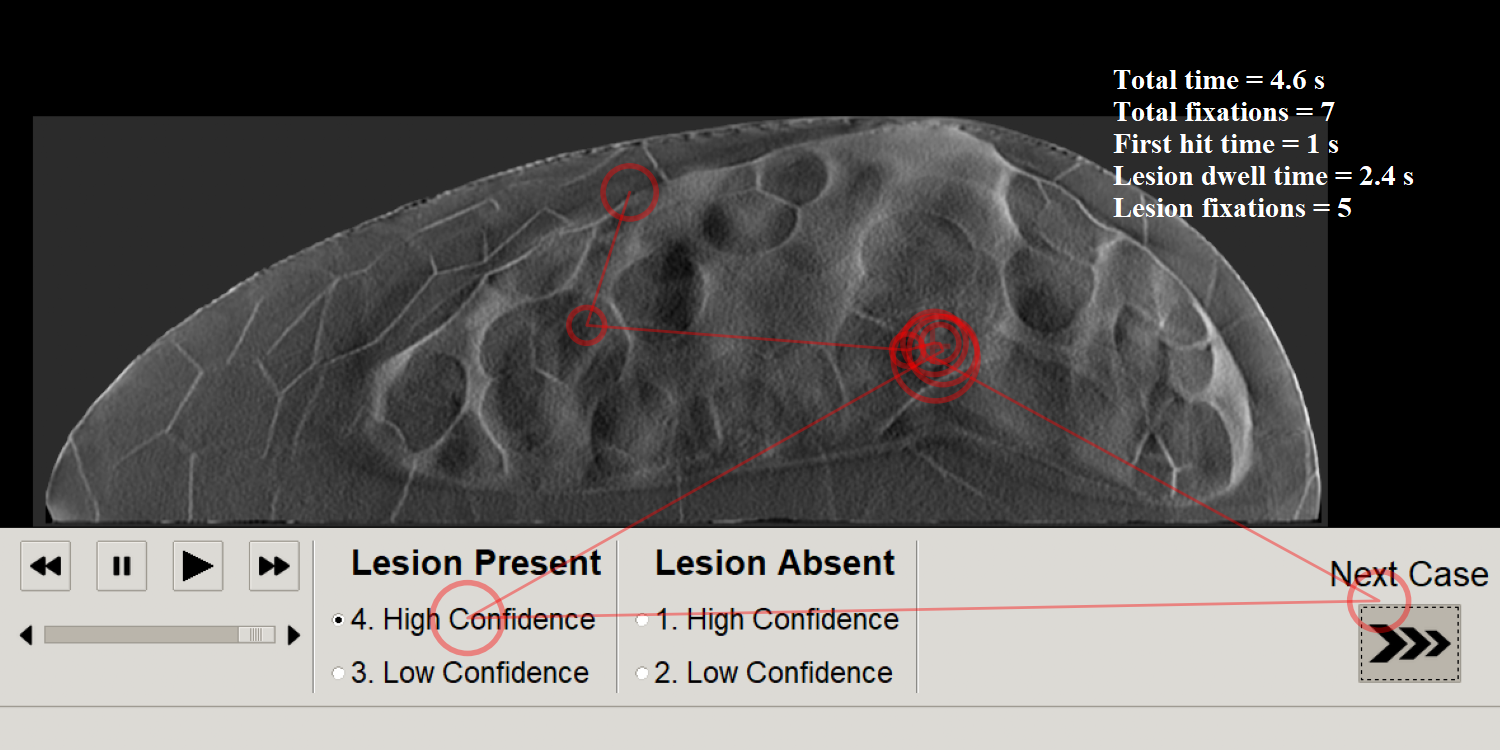}
         \caption{\label{fig:pd25}25\% dense Bakic background}
     \end{subfigure}
\vskip\baselineskip
     \begin{subfigure}[b]{0.6\textwidth}
         \centering\includegraphics[width=\textwidth]{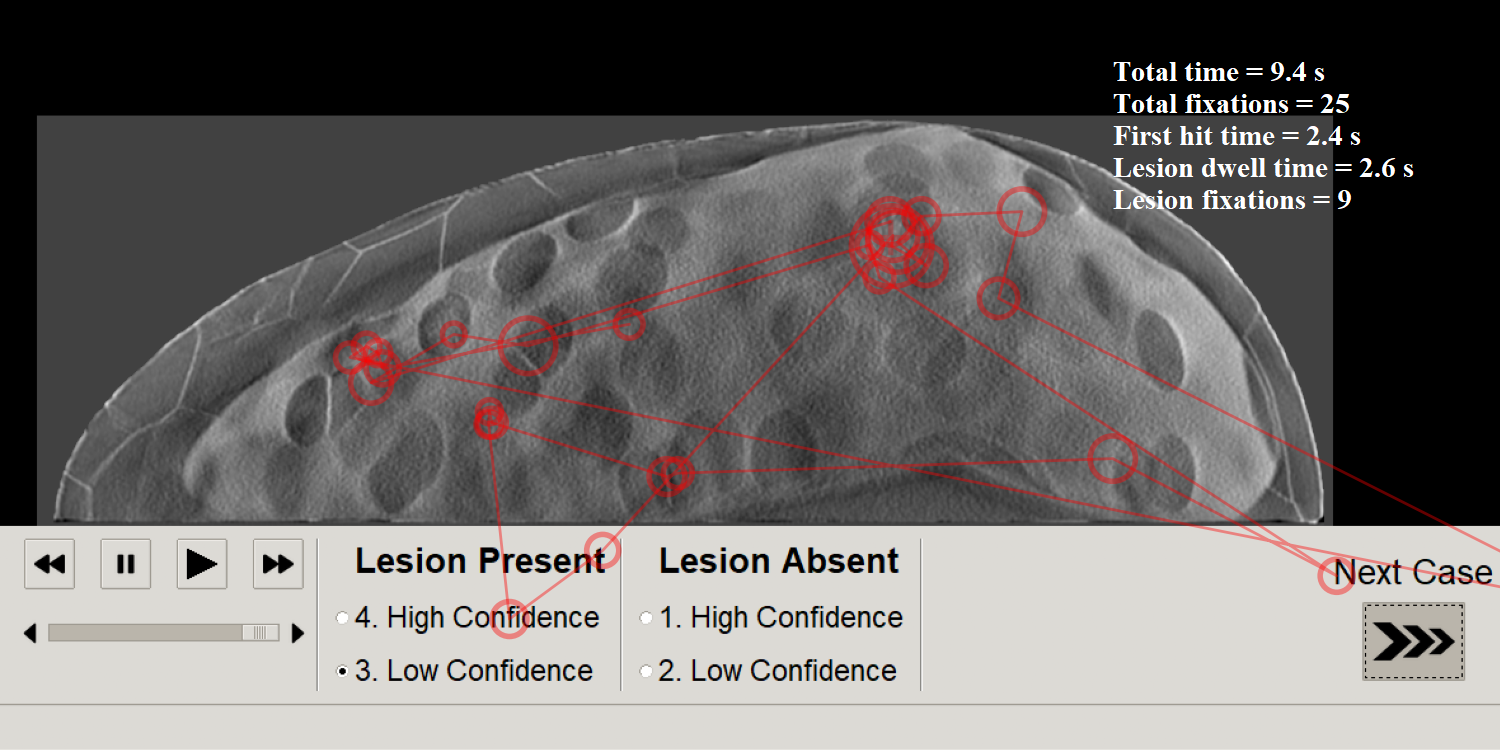}
         \caption{\label{fig:pd50}50\% dense Bakic background}
     \end{subfigure}

        \caption{\label{fig:img_fixations}Differences between search patterns of an observer on an DBT slice of a 25\% dense XCAT breast phantom (a), a 25\% dense Bakic phantom (b), and a 50\% dense  Bakic phantom (c). The size of the circles are proportional to the fixation duration. Observer made fewer fixations and diagnosed quickly 25\% dense and XCAT breast images than 50\% dense and Bakic images respectively.}
\end{figure}

Heterogeneity in patient anatomies can influence the diagnostic performance and interpretation process.
 In this study, heterogeneity was achieved by simulating DBT images with digital phantoms of different breast densities (25\% and 50\% dense) and structures (Bakic and XCAT breast phantoms). 
We collected gaze data for in-plane DBT images for the task of searching and locating a spicule and sphere lesions.
 Figure \ref{fig:img_fixations} shows a sample gaze pattern differences due to change in breast density and structures.
% [Observers made fewer fixations and diagnosed quickly on images of breast with 25\% density than with 50\% density and XCAT breast backgrounds than Bakic backgrounds.] 

\begin{figure}[ht!]
     \centering
    \begin{subfigure}[b]{0.29\textwidth}
         \centering\includegraphics[width=\textwidth]{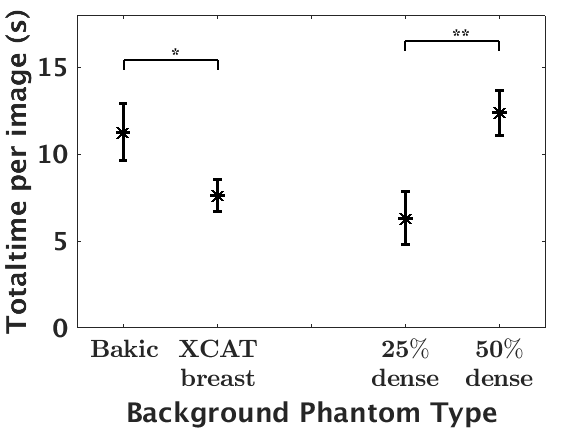}
         \caption{\label{fig:Ttime_pht}}
     \end{subfigure}
     \hfill 
    \begin{subfigure}[b]{0.29\textwidth}
         \centering\includegraphics[width=\textwidth]{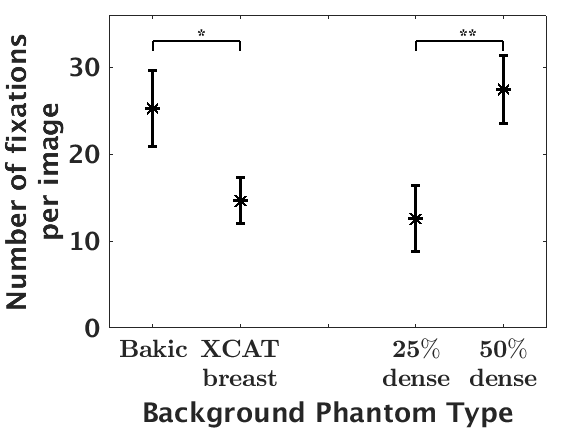}
         \caption{\label{fig:Nfix_pht}}
     \end{subfigure}
     \hfill
    \begin{subfigure}[b]{0.29\textwidth}
         \centering\includegraphics[width=\textwidth]{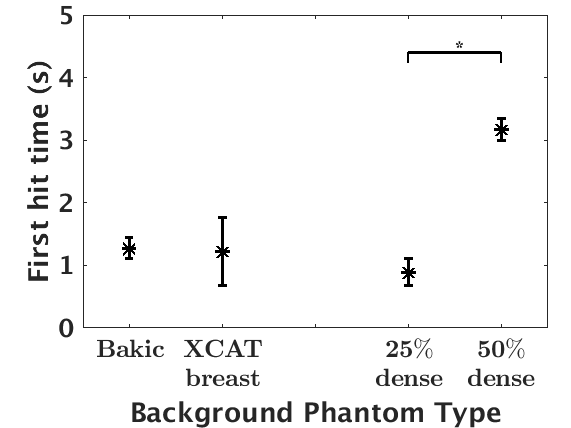}
         \caption{\label{fig:FHT_pht}}
     \end{subfigure}
\vskip\baselineskip
     \hfill
     \begin{subfigure}[b]{0.29\textwidth}
         \centering\includegraphics[width=\textwidth]{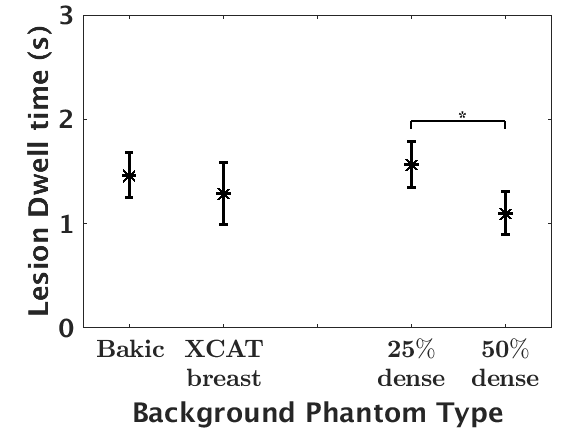}
         \caption{\label{fig:Ltime_pht}}
     \end{subfigure}
     \hfill
     \begin{subfigure}[b]{0.29\textwidth}
         \centering\includegraphics[width=\textwidth]{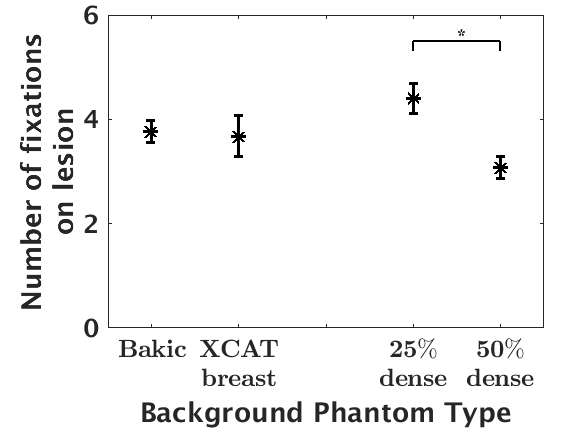}
         \caption{\label{fig:Lfix_pht}}
     \end{subfigure}
     \hfill
        \caption{\label{fig:Gze_pht}The average amount of time spent and number of fixations made on each image, first hit time, lesion dwell time, and  number of fixations on lesion were plotted for different breast densities and phantom types. Error bar lengths indicate twice the standard error of the six observers’ average gaze metrics. Observers made less number of fixations and diagnosed quickly 25\% dense images than 50\% dense images and XCAT breast backgrounds than Bakic backgrounds. Observers took longer to first fixate on lesion in 50\% dense images than 25\% dense images whereas no difference was observed due to change of phantom type. Observers made fewer fixations on lesion when inserted in 50\% dense images than in 25\% dense images whereas no difference was observed due to change of phantom type. * represents a statistical significant difference with 0.01< p-value < 0.05. ** represents a statistical significant difference with p-value < 0.01.}
\end{figure}

Figure \ref{fig:Gze_pht} shows the average gaze metrics of the six observers for different densities and phantom backgrounds. 
Error bar lengths indicate twice the standard error of the average gaze metrics for the six observers. 
Variation in breast density and phantom type lead to significant differences in the total time and number of fixations made by each observer.
Overall, the time needed to reach a diagnosis was shorter for images of breast with 25\% density relative to 50\% density (p-value < 0.01) and for the XCAT breast phantom backgrounds compared to Bakic backgrounds (p-value < 0.05).
In addition, observers made fewer fixations on images of breast with 25\% density relative to 50\% density (p-value < 0.01) and on XCAT breast phantom backgrounds compared to Bakic backgrounds (p-value < 0.05). 
The amount of time required to first fixate on the lesion (first hit time) was larger in the images with higher density (p-value < 0.05) whereas no difference was observed due to change of phantom background. 
Similarly, lesion dwell time and lesion fixations were insensitive to changes in the type of phantom background. 
However, observers on average spent less time (p-value < 0.01) and made fewer fixations (p-value < 0.01) on the lesion in higher dense images. This is because observers failed to fixate on more number of lesions when looking at higher density images.

%\subsection{Effect of signal size and type}

\begin{figure}[ht!]
     \centering
    \begin{subfigure}[b]{0.29\textwidth}
         \centering\includegraphics[width=\textwidth]{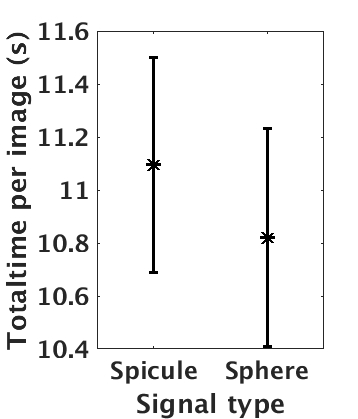}
         \caption{\label{fig:Ttime_signals}}
     \end{subfigure}
     \hfill 
    \begin{subfigure}[b]{0.29\textwidth}
         \centering\includegraphics[width=\textwidth]{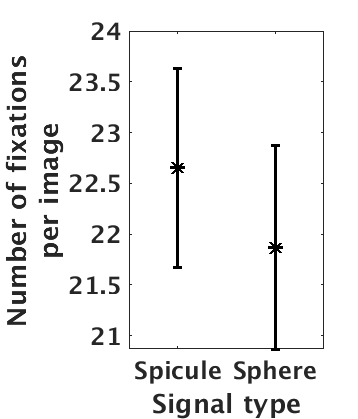}
         \caption{\label{fig:Nfix_signals}}
     \end{subfigure}
     \hfill 
     \begin{subfigure}[b]{0.29\textwidth}
         \centering\includegraphics[width=\textwidth]{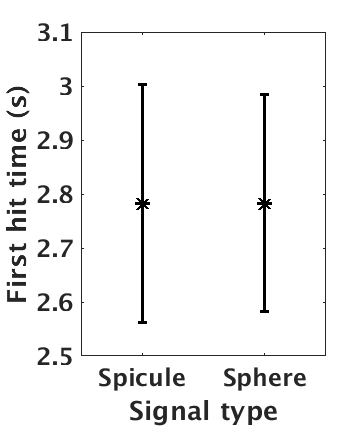}
         \caption{\label{fig:FHT_signals}}
     \end{subfigure}
\vskip\baselineskip
     \hfill
     \begin{subfigure}[b]{0.29\textwidth}
         \centering\includegraphics[width=\textwidth]{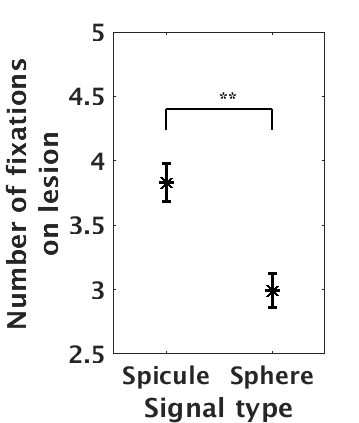}
         \caption{\label{fig:Lfix_signals}}
     \end{subfigure}
    \begin{subfigure}[b]{0.29\textwidth}
         \centering\includegraphics[width=\textwidth]{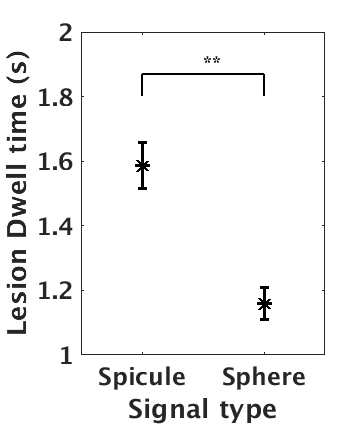}
         \caption{\label{fig:Ltime_signals}}
     \end{subfigure}
     \hfill 
        \caption{\label{fig:Gze_signals}The average amount of time spent on each image, number of fixation per image, first hit time, lesion dwell time, and number of fixations on lesion were plotted for change of signal type. Observers spent less time and made lesser number of fixations on sphere lesion compared to spicule lesion. No significant differences in other gaze metrics observed due to change of signal type.}
\end{figure}

The target signal size and type are other factors that, like anatomical structures, can influence the interpretation process. 
%We conducted human observer study for the task of searching and locating sphere and spicule signals in two separate sessions under similar system and phantom configurations and study conditions. Using the eye-tracking data, we computed the five gaze metrics and the three error signatures. 
Figure \ref{fig:Gze_signals} shows the average gaze metrics of the six observers for two signal types. Error bar lengths indicate twice the standard error of the average gaze metrics for the six observers. Observers spent less time (p-value < 0.01) and made lesser number of fixations (p-value <0.01) on sphere lesion compared to spicule lesion. No significant differences were found in other gaze metrics due to change of signal type.
%Observers needed more time to first fixate and to localize the spherical lesion, whereas observers spent more time and made more fixations on the spiculated lesion. These differences were not statistically significant. 
\subsection{Attention analysis}
\begin{figure}[ht!]
     \centering
    \begin{subfigure}[b]{0.9\textwidth}
         \centering\includegraphics[width=0.9\textwidth]{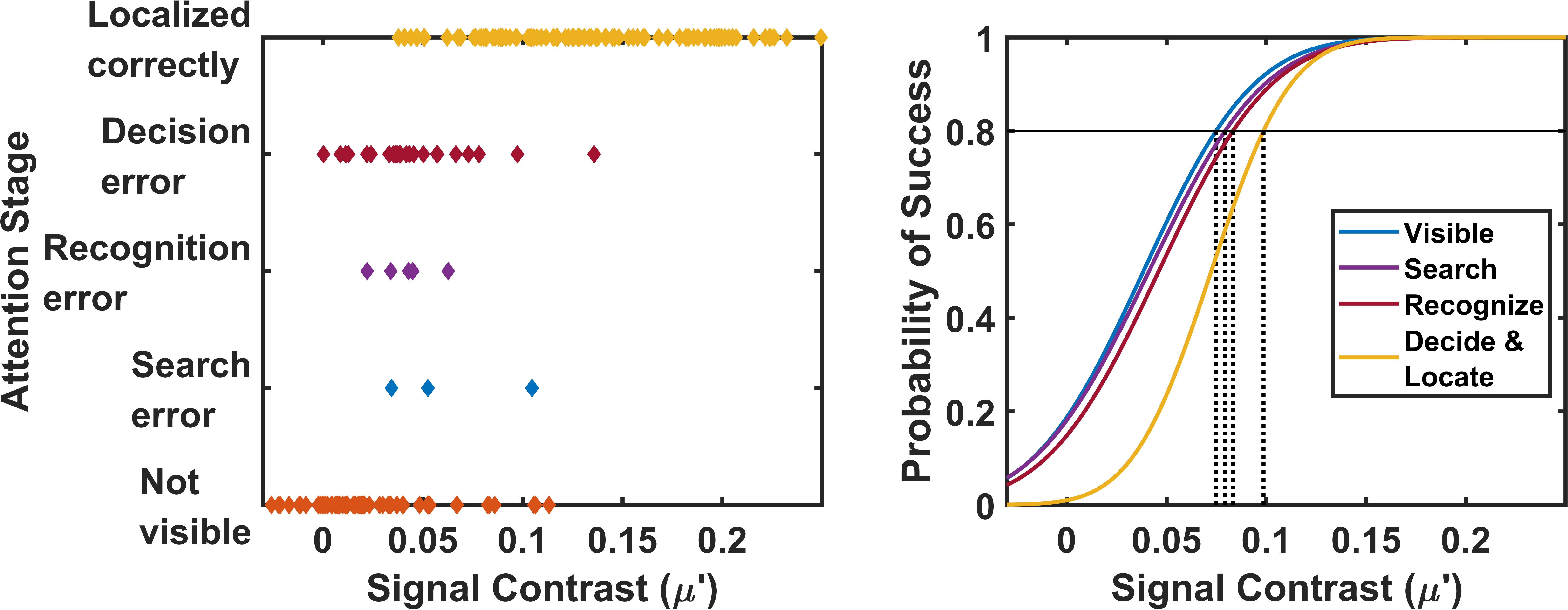}
        % \caption{\label{fig:attn_pht}}
     \end{subfigure}
\hfill
     \begin{subfigure}[b]{0.9\textwidth}
         \centering\includegraphics[width=0.9\textwidth]{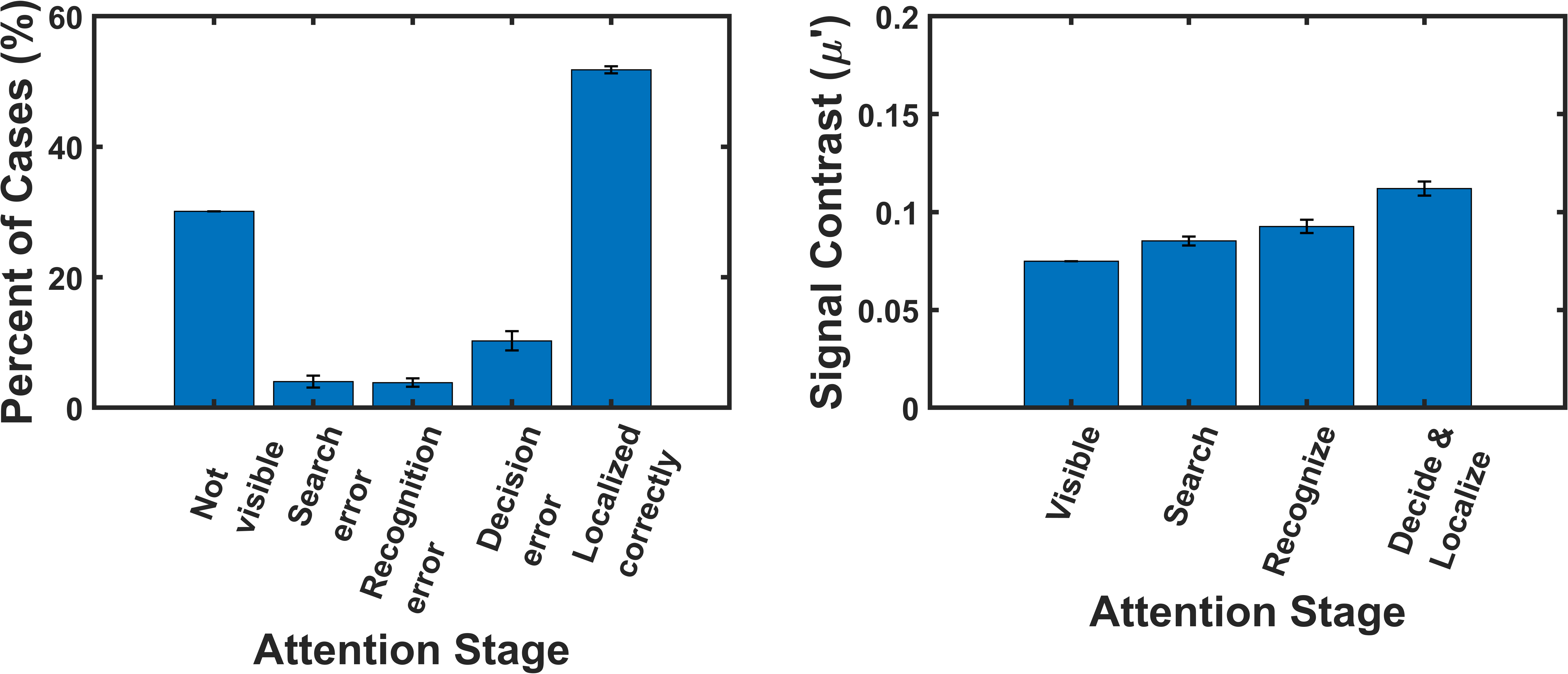}
         %\caption{\label{fig:attn_vgf}}
     \end{subfigure}
\caption{\label{fig:attn_ant}The cases that fall in each attention stage for an observer plotted against signal contrast on the top left and the average percent of cases fall in each attention stage on the bottom left. Sample gaussian cdf fits for signal contrast in each attention stage on the top right and the average required signal contrast of the six observers for each attention stage. Highest number of cases were misdiagnosed (FN) due to signals were not visible followed by decision errors. 
Significantly higher signal contrast required at the decision stage than that of recognition stage. }
\end{figure}

We categorized each case into a distinct attention category — not visible, search error, recognition error, decision error, and localized correctly—for each observer independently. Figure \ref{fig:attn_ant} (top left) presents the distribution of cases across these attention categories as a function of signal contrast of an observer, while the bottom left panel shows the average percentage of cases falling into each category across all observers.
The highest proportion of misdiagnosed cases (false negatives) occurred when the signal was not visible, accounting for 25\% of total positive cases. This was followed by errors in the decision phase (11\%), where the signal was seen and recognized but an incorrect judgment was made. Recognition and search-stage errors each accounted for 4\% of misdiagnoses, indicating that while relatively rare, errors at all stages can contribute to diagnostic failure.
To estimate the signal contrast thresholds required for successful completion of each attention stage, we fit cumulative Gaussian functions to the data for each observer and attention stage. The average goodness-of-fit ($R^2$) across all fits was $0.89 \pm 0.06$, indicating a strong correspondence between observed and modeled data. The fitted psychometric curves (Figure \ref{fig:attn_ant}, top right) were used to compute the contrast level required to achieve an 80\% success rate in each stage. The bottom right panel shows the average 80\%-threshold contrast across six observers. 
We found that the decision stage required significantly higher signal contrast than the recognition stage. Specifically, the average contrast required for 80\% success in the decision stage was $0.096\pm0.002$, compared to $0.078\pm0.003$ for the recognition stage and  $0.072\pm0.002$ for the search stage —a statistically significant difference (paired t-test, p<0.01), indicating that location unknown constraints primarily affect later attention stages.

\begin{figure}[ht!]
     \centering
    \begin{subfigure}[b]{0.45\textwidth}
         \centering\includegraphics[width=\textwidth]{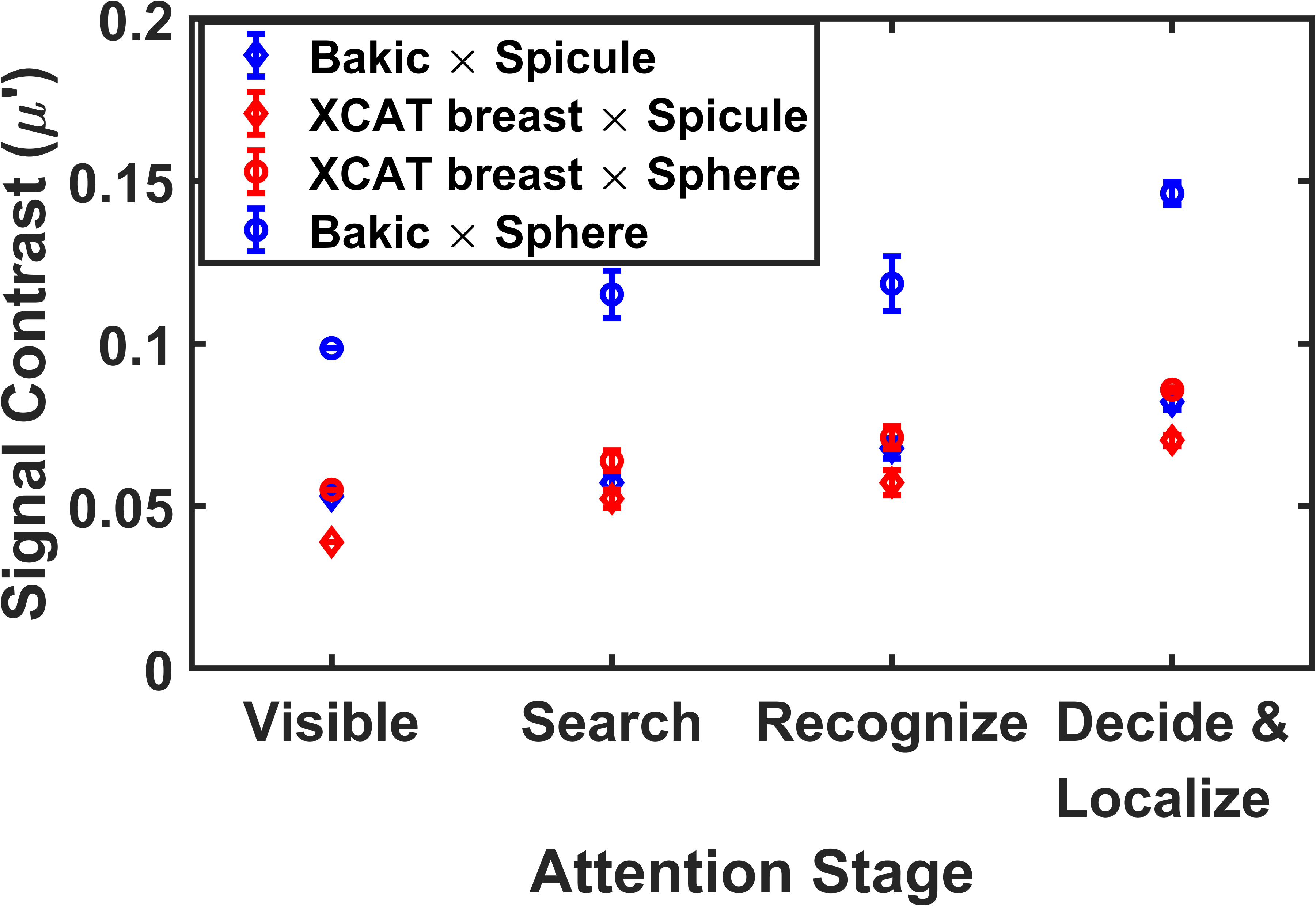}
        % \caption{\label{fig:attn_ctrst1}}
     \end{subfigure}
\caption{\label{fig:attn_ctrst}The signal contrast required for each attention stage plotted for two background types and signal types. The length of error bars represent the twice the standard error of the six observers.}
\end{figure}

Figure \ref{fig:attn_ctrst} breaks down further the contrast thresholds by lesion type (spiculated vs. spherical) and background complexity (XCAT vs. Bakic breast phantoms). For spiculated lesions, attentional-stage performance was relatively similar across both background types, suggesting that the local structure of spiculations confers sufficient salience even in cluttered anatomical noise. In contrast, spherical lesions—particularly those embedded in the Bakic backgrounds—showed a marked increase in required contrast across all attention stages.
Interestingly, in the XCAT backgrounds, both lesion types yielded similar contrast thresholds across attention stages, indicating that scene complexity plays a key role in mediating attention-related perceptual errors. Within the Bakic background, however, spherical lesions consistently required higher contrast than spiculated lesions for search, recognition, and decision, highlighting the interaction between target structure and visual clutter in perceptual processing.

% The contrast requirement for decision to localize a signal is significantly higher than that of recognition stage. 

\section{Discussion}

We evaluated the differences in human gaze metrics and error signatures to understand how signal and background influence different visual attention mechanisms during the diagnostic task.
The results demonstrate that both background complexity and lesion structure significantly modulate observer behavior and perceptual performance in digital breast imaging. Observers took longer to interpret high-density (50\%) breast images and those generated from Bakic phantoms, and they made more fixations in these conditions compared to lower-density (25\%) or XCAT-generated images. These effects reflect increased visual clutter and structural heterogeneity in denser or more complex anatomical scenes, which can hinder target salience and reduce the efficiency of both pre-attentive and focal search mechanisms.
However, once the lesion was fixated, dwell time and number of fixations were not significantly impacted by background structure alone, indicating that post-detection processing may proceed relatively uniformly across background types—assuming the signal is acquired in the first place.

Conversely, lesion type had a clear impact on gaze behavior: spiculated lesions attracted longer fixations and more attention than spherical lesions, likely due to their distinct edge structure and higher local contrast gradients, which are known to guide visual attention more effectively. These differences in gaze behavior align with established findings in visual search literature, where texture and orientation discontinuities increase bottom-up salience.

Error signature analysis further deepens this understanding by highlighting how signal visibility, search, recognition, and decision stages independently contribute to diagnostic performance. A striking 25\% of all positive cases were missed due to complete signal invisibility—underscoring the impact of low detectability, especially in dense breast regions. Even when signals were visible, the decision stage required significantly higher contrast thresholds (0.096 ± 0.002) than the recognition stage (0.078 ± 0.003), suggesting that perceptual ambiguity or internal criterion variability disproportionately impacts decision-level processing. This aligns with perceptual decision-making frameworks that emphasize the compounding of uncertainty at later cognitive stages, where prior knowledge and confidence thresholds begin to dominate

When stratifying results by lesion type and background, a clear interaction emerged. While spiculated lesions were relatively robust to changes in anatomical complexity, spherical lesions, especially in Bakic backgrounds, demanded substantially higher contrast to reach equivalent success rates across all attention stages. This confirms that lesion geometry and contextual integration with surrounding anatomy jointly determine detectability. The smaller spherical lesion may be masked by high frequency structures of Bakic backgrounds.

Together, these findings highlight the multifaceted nature of perceptual challenges in medical image interpretation. Performance is not solely determined by signal strength but emerges from a complex interaction between observer attention, lesion salience, and background structure. These insights reinforce the need for diagnostic systems and training protocols that explicitly address perceptual limitations at each stage of visual processing. Enhancing lesion conspicuity through contrast optimization, background suppression, or gaze-guided decision aids could substantially improve outcomes, especially in dense breast imaging contexts where perceptual bottlenecks are most pronounced.

\section{Acknowledgments}
This work was partially supported by funding from the NIH National Institute of Biomedical Imaging and
Bioengineering (NIBIB) granta R01 EB EB029761 and R01EB032416 , the US Department of Defense (DOD) Congressionally Directed Medical Research Program (CDMRP) Breakthrough Award BC151607 and the National Science Foundation CAREER Award 1652892. We would also like to thank both UPenn and Duke groups for making the digital phantoms freely available for research. 

\bibliography{references}
\bibliographystyle{unsrt}

\end{document}